\pdfoutput=1

\documentclass[11pt]{article}

\usepackage[final]{acl}

\usepackage{times}
\usepackage{latexsym}
\usepackage{amssymb}
\usepackage[T1]{fontenc}
\usepackage{soul}
\usepackage{xcolor}
\usepackage[utf8]{inputenc}
\usepackage{amsmath}
\usepackage{microtype}
\usepackage{booktabs}
\usepackage{inconsolata}
\usepackage{enumitem}
\usepackage{tabularx}
\usepackage{graphicx}
\usepackage{makecell}
\usepackage[ruled,linesnumbered]{algorithm2e}
\usepackage{multirow}
\usepackage{bbding}
\usepackage{hyperref}
\title{Continual Gradient Low-Rank Projection Fine-Tuning for LLMs}

\author{{\bf Chenxu Wang\textsuperscript{1}},  {\bf Yilin Lyu\textsuperscript{1}},  {\bf Zicheng Sun\textsuperscript{1}},  {\bf Liping Jing\textsuperscript{1}\thanks{Corresponding authors.}} \\
    \textsuperscript{1}Beijing Key Laboratory of Traffic Data Mining and Embodied Intelligence \\
    School of Computer Science and Technology, Beijing Jiaotong University \\
    State Key Laboratory of Advanced Rail Autonomous Operation \\ 
    \texttt{\{chenxuwang, yilinlyu, zichengsun, lpjing\}@bjtu.edu.cn}}

\begin{document}
\maketitle
\begin{abstract}

Continual fine-tuning of Large Language Models (LLMs) is hampered by the trade-off between efficiency and expressiveness. Low-Rank Adaptation (LoRA) offers efficiency but constrains the model's ability to learn new tasks and transfer knowledge due to its low-rank nature and reliance on explicit parameter constraints. 
We propose GORP (\underline{\textbf{G}}radient L\underline{\textbf{O}}w \underline{\textbf{R}}ank \underline{\textbf{P}}rojection) for Continual Learning, a novel training strategy that overcomes these limitations by synergistically combining full and low-rank parameters and jointly updating within a unified low-rank gradient subspace. GORP expands the optimization space while preserving efficiency and mitigating catastrophic forgetting. 
Extensive experiments on continual learning benchmarks demonstrate GORP's superior performance compared to existing state-of-the-art approaches. Code is available at~\href{https://github.com/Wcxwcxw/GORP}{https://github.com/Wcxwcxw/GORP}.

\end{abstract}

\section{Introduction}

Large Language Models (LLMs) have demonstrated remarkable capabilities in areas like in-context learning \citep{hendel-etal-2023-context, pmlr-v235-liu24bx} and instruction following \citep{DBLP:conf/nips/Wei0SBIXCLZ22, wei2022emergent}.
To adapt these large models to specific downstream tasks, traditional full fine-tuning imposes prohibitive computational costs and memory requirements, which has driven extensive research into parameter-efficient fine-tuning (PEFT) approaches \citep{pmlr-v97-houlsby19a,hu2022lora,ben-zaken-etal-2022-bitfit}.
Low-Rank Adaptation (LoRA) \citep{hu2022lora}, in particular, has become a popular PEFT technique, especially in continual learning scenarios \citep{chitale2023task, wistuba2024continual}, due to its efficiency and ability to mitigate catastrophic forgetting \citep{biderman2024lora}.


While LoRA significantly reduces training complexity and storage, the low-rank matrices inherently constrain the parameter space and, consequently, the model's expressiveness during optimization \citep{DBLP:conf/icml/Zhao0CWAT24}. This restriction to a low-rank subspace can lead to suboptimal performance compared to full fine-tuning, a gap that often widens in continual learning settings \citep{xia2024chainloraefficientfinetuning,mahla2025exploringgradientsubspacesaddressing}. Furthermore, LoRA updates are intertwined with shared parameter updates, potentially causing collisions in the parameter spaces of different tasks \citep{wang-etal-2023-orthogonal,lu2024controlledlowrankadaptationsubspace}.
Gradient projection has emerged as a promising mitigation strategy \citep{saha2021gradient,Wang_2021_CVPR,Kong_balancingstability,Saha_Roy_2023}. Common approaches involve calculating the hidden feature space and projecting it onto the orthogonal gradient space of the old task. However, gradient spaces for different tasks are heterogeneous and dynamically evolving. Existing methods that impose explicit constraints (e.g., parameter regularization) on LoRA's low-rank parameters \citep{wang-etal-2023-orthogonal,du-etal-2024-unlocking,yang-etal-2025-parameter} can only approximate the ideal parameter space and fail to adapt dynamically to the changing gradient space of new tasks \citep{liu2024learningattentionalmixtureloras}. Moreover, these explicit constraints often struggle to capture shared features across tasks, hindering knowledge transfer.

\begin{table*}
    \centering
    \scalebox{0.9}{
    \begin{tabular}{lcccccc}
    \toprule
        & \multicolumn{2}{c}{\textbf{Parameters}} & \multicolumn{2}{c}{\textbf{Parameter Constraints}} & \multicolumn{2}{c}{\textbf{Gradient Space}}  \\
    \textbf{Method} & \textbf{Full-rank}  & \textbf{Low-rank} & \textbf{Explicit} & \textbf{Implicit} & \textbf{Low-rank} & \textbf{Adaptability} \\
    \cmidrule(r){1-1} \cmidrule(r){2-3}  \cmidrule(r){4-5} \cmidrule(r){6-7}
    O-LoRA \citep{wang-etal-2023-orthogonal} & \XSolidBrush & \Checkmark & \Checkmark & \XSolidBrush & \XSolidBrush & Static\\
    MIGU \citep{du-etal-2024-unlocking} & \XSolidBrush & \Checkmark   & \XSolidBrush & \Checkmark & \XSolidBrush &  Static\\
    N-LoRA \citep{yang-etal-2025-parameter} & \XSolidBrush & \Checkmark  & \Checkmark & \XSolidBrush & \XSolidBrush & Static\\
    \cmidrule(r){1-1} \cmidrule(r){2-3}  \cmidrule(r){4-5} \cmidrule(r){6-7}
    GORP(Ours) & \Checkmark & \Checkmark  & \XSolidBrush & \Checkmark & \Checkmark & Dynamic\\
    \bottomrule
    \end{tabular}}
    \caption{Comparison of continual fine-tuning methods on training parameters, parameter constraints and Gradient Space Adaptability.}
    \label{tab:method_comparison}
\end{table*}

To address these limitations, we introduce GORP (\underline{\textbf{G}}radient L\underline{\textbf{O}}w \underline{\textbf{R}}ank \underline{\textbf{P}}rojection) for Continual Learning, a novel training strategy for continual fine-tuning of LLMs that synergistically integrates full and low-rank parameter updates within a low-rank gradient subspace. GORP effectively balances the \emph{stability-plasticity dilemma} inherent in continual learning (see Table~\ref{tab:method_comparison} for a comparison with other methods). From a \emph{plasticity} perspective, GORP enhances LoRA by incorporating learnable full-rank parameters for the current task. Crucially, we exploit the observation that gradients tend to adopt a low-rank structure during training — a phenomenon theoretically supported and broadly observed in neural architectures like transformers
\citep{DBLP:conf/icml/Zhao0CWAT24}. Therefore, we project the gradients of these full-rank parameters into a low-rank space, maintaining fine-tuning efficiency while significantly expanding the search space for optimal solutions.
From a \emph{stability} perspective, GORP departs from prior methods that rely on explicit constraints. Recognizing the limitations of directly sampling subspaces from large-scale models, we leverage the first-order moment of gradients to implicitly capture the dynamic properties of the gradient space. This approach provides a more robust and comprehensive representation of the gradient, reducing computational complexity compared to methods that directly manipulate the hidden feature space \citep{saha2021gradient,zheng2024antiforgettingmultimodalcontinualinstruction,qiao2024gradientprojectioncontinualparameterefficient}.
We evaluate GORP on several continual fine-tuning evaluations, demonstrating its superior performance compared to existing state-of-the-art methods. Our results confirm that GORP provides a more effective approach for continual fine-tuning of LLMs.

Our main contributions are summarized as follows:
\begin{itemize}[leftmargin=*]
    \item We leverage the complementary strengths of full and low-rank parameters by jointly updating them within a unified low-rank gradient subspace. This expands the search space for optimal solutions while retaining the efficiency of low-rank adaptation.
    \item We utilize the first-order moment of gradients to approximate the hidden feature space, providing a more robust and efficient way to construct a gradient subspace. This mitigates catastrophic forgetting and minimizes computational overhead.
    \item We introduce GORP, a novel training strategy that effectively balances stability and plasticity in continual learning, outperforming existing methods while maintaining fine-tuning efficiency.
\end{itemize}

\section{Related Works}
\subsection{Parameter-efficient Fine Tuning of LLMs}
Various efficient parameter fine-tuning methods include adapters \citep{pmlr-v97-houlsby19a}, Low-Rank Adaptation (LoRA) \citep{hu2022lora}, and parameter subset techniques \citep{ben-zaken-etal-2022-bitfit}. These methods have tackled the challenges including large number of parameters and substantial memory requirements by fine-tuning selective model parameters rather than the entire model. Among these, LoRA has become one of the most widely used methods, which is achieved by freezing pre-trained weights and introducing low-rank trainable matrices, effectively reducing the computational burden. Building on LoRA,  \citet{lialin2023relorahighranktraininglowrank} proposed a series of low-rank aggregation updates for learning network parameters. \citet{xia2024chainloraefficientfinetuning} employed a residual LoRA module at each fixed step, and eventually merging it with the pre-trained model parameters for chained updates. \citet{hao2024flora} used random projection sampling to approximate LoRA, enabling high-rank weight updates, and optimizing memory usage.

\subsection{Continual Fine Tuning for LLMs}
Three widely used continual learning paradigms \citep{shi2024continuallearninglargelanguage,lu2024controlledlowrankadaptationsubspace,zheng2024lifelonglearninglargelanguage} for parameter fine-tuning are Replay-based methods \citep{zhao-etal-2022-prompt,huang-etal-2024-mitigating}, Architecture-based methods \citep{badola-etal-2023-parameter,song2023conpetcontinualparameterefficienttuning}, and Learning-based methods \citep{pmlr-v108-farajtabar20a,DBLP:journals/tmlr/SmithHZHKSJ24}, which employ specific optimization strategies or introduce regularization penalties based on the original loss function to balance the trade-off between old and new knowledge.
Many studies have demonstrated improved performance through learning-based methods. 
\citet{qiao2024gradientprojectioncontinualparameterefficient} proposed an overarching framework for continual fine-tuning, establishing diverse paradigms for efficient fine-tuning. However, due to the challenges in obtaining gradient spaces and the impracticality of using implicit feature spaces, \citet{wang-etal-2023-orthogonal} suggested leveraging LoRA itself to represent the gradient space, ensuring orthogonality between gradient spaces of different tasks to mitigate forgetting. Subsequently, \citet{du-etal-2024-unlocking} focused on screening the normalized gradients of the hidden linear layer outputs and updating the selected parameters to minimize gradient conflicts. \citet{yang-etal-2025-parameter} introduced parameter sparsification constraints, addressing parameter conflicts between tasks and ensuring that each task’s vector space remains independent. Additionally, \citet{lu2024controlledlowrankadaptationsubspace} and \citet{chen2024bayesianparameterefficientfinetuningovercoming} employed regularization matrices and introduced further constraints to enhance the ability of LLMs to learn new tasks.

\subsection{Continual Learning with Gradient Projection}
Gradient projection methods in continual learning project the gradient into a subspace of the input's implicit feature space to mitigate catastrophic forgetting when learning new tasks. The Gradient Projection Memory proposed by \citet{saha2021gradient} leverages the relationship between the input and gradient spaces to form a gradient subspace for each layer, thereby retaining prior knowledge while accommodating new information.
However, the gradient space can impose restrictive constraints on the optimization space for new tasks, potentially limiting their learning performance. To facilitate both forward and backward knowledge transfer, \citet{DBLP:conf/iclr/LinYFZ22}\citeyearpar{NEURIPS2022_6728fcf9} proposed a scaling matrix based on the similarity between new and previous tasks, using the frozen weights from the old task to scale and update the current task’s weights. In response to the continuous expansion of the gradient subspace, \citet{10204088} introduced the dual gradient projection memory method, which reduces memory consumption and adaptively expands the dimensionality of the layer, enhancing the model's plasticity for new tasks. Other studies \citep{Kong_balancingstability,Wang_2021_CVPR,Lin_2022_CVPR} also improved continual learning performance by refining the gradient space.

\begin{figure*}[t]
  \centering
  \includegraphics[width=\linewidth]{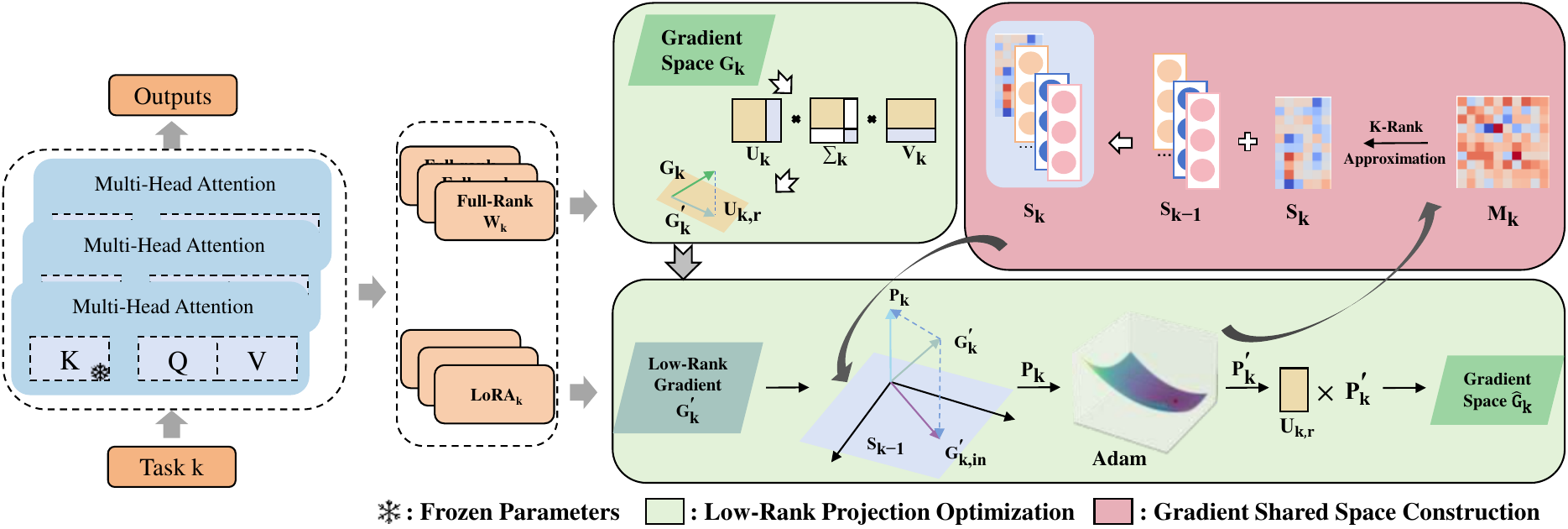}
  \caption {The framework of our Gradient Low Rank Projection (GORP) method. During $k$-th task training, we reduce the dimensions of full-rank parameters and project both full and low-rank parameters into the space $\mathcal{S}_{k-1}$. Then, we use the first-order moment $M_k$ and a k-rank approximation to construct the Gradient Shared Space $\mathcal{S}_k$.}
  \label{fig:framwork}
\end{figure*}

\section{Gradient Low Rank Projection}
We introduce GORP, a novel training strategy that combines full and low-rank parameters with low-rank gradient updates to strike a balance between plasticity and stability. The framework, illustrated in Figure~\ref{fig:framwork}, consists of two main components: (1) the Gradient Shared Space Construction, which employs low-rank moment with distinct parameters to construct a shared gradient space, and (2) the Low-Rank Projection Optimization, which projects the gradient space of both full and low-rank parameters. The pseudo-code of our method is provided in Algorithm~\ref{algorithm}.

\begin{algorithm}[t]
\caption{GORP}
\label{algorithm}
\SetKwInOut{Input}{Input}\SetKwInOut{Output}{Output}
\SetKwFunction{Project}{Project}
\Input{Old task weight $W$, gradient $G_t$, step $t$, rank $r$, scale factor $\alpha$, decay rates $\beta_1,\beta_2$, learning rate $\eta$, subspace change frequency $T$, num steps $N$.}
\Output{New task weight $W$}
Initialize gradient subspace $\mathcal{S}\leftarrow [\;]$\\
Initialize first-order moment $M_t \leftarrow 0$\\
Initialize second-order moment $V_t \leftarrow 0$\\
Initialize step $t \leftarrow 1$\\
\While{$t \leq N$}
{
    \If{Full-rank Parameters}{ 
        \eIf(\tcp*[h]{via Equation~\ref{eq:Gt_svd}}){$t\;\mathrm{mod}\;T=0$}
        {$USV \leftarrow \mathrm{SVD}(G_t)$\\
        $G_t^{'} \leftarrow U_r^{\top}G_t$}
        {$G_t^{'} \rightarrow G_{t-1}^{'}$}
        }
    \If{LoRA Parameters}{
        $G_t^{'} \leftarrow G_t$\\
    }
    $P_t\leftarrow \Project(G_t^{'})$\tcp*[h]{via Equation~\ref{eq:Gt_svd0}}\\
    $M_t \leftarrow \beta_1 M_{t-1} + (1-\beta_1) P_{t}$\\
    $V_t \leftarrow \beta_2 V_{t-1} + (1-\beta_2) P_{t}^2$\\
    $P_t^{'} \leftarrow M_{t}/\sqrt{V_{t}+\epsilon}$\\
    $W_t \leftarrow W_{t-1} + \eta\cdot\alpha U_rP_t^{'}$
}
$\mathrm{Update} \; \mathcal{S} \; \mathrm{with} \; M_t$\tcp*[h]{via Equation~\ref{eq:grad_proj},\ref{eq:secondtask_k},\ref{eq:update_S}}\\
\Return{New task weight $W$}
\end{algorithm}

\subsection{Gradient Shared Space Construction}
In this section, we construct a gradient shared space. A common approach for building gradient spaces in continual learning is to randomly sample from hidden layer input features. However, for LLMs trained on vast amounts of data, the limited number of sampled features may fail to accurately represent the overall data distribution. Consequently, the resulting gradients may not align with the overall gradient direction during gradient space computation.

To address this issue, we employ low rank moment to more accurately represent the overall gradient space. Specifically, using Adam as an example, for the parameter gradient $G_t\in \mathbb{R}^{m \times n}$, there exists a first-order moment $M_t \in \mathbb{R}^{m \times n}$.
Since Adam incorporates historical gradient information at each iteration, its moment term can theoretically help the optimization algorithm better approximate the optimal gradient direction for the overall task, particularly when the task's loss function exhibits a flat or irregular landscape. Thus, after training, we can leverage first-order moment information to capture the gradient direction of the current task and calculate the gradient sharing space. Let L denote the number of parameter layers to be trained.

For the first task, we utilize first-order moments of each layer's parameters, denoted as $M_1 = \{M_1^1, M_1^2,\dots, M_1^l, \dots, M_1^L\}$. We then perform singular value decomposition (SVD) on each layer, yielding $M_1^l = U_1^l\sum_1^l{V_1^l}^{\top}$. Finally we execute a k-rank approximation under the specified constraints:
\begin{equation}
  \label{eq:firsttask_k}
  \|(M_1^l)_k\|_F^2 > \epsilon_{t}^l \|M_1^l\|_F^2
\end{equation}
where $\epsilon_{t}^l$ is an approximation threshold. We select the first k vectors from $U_1^l$ to form layer gradient space, denoted as $\mathcal{S}_1^l = [ u_{1,1}^l, u_{1,2}^l,\dots,u_{1,k}^l]$, and aggregate the layer-wise gradient spaces to obtain overall gradient space $\mathcal{S} = \{\{\mathcal{S}_1^l\}_{l=1}^L\}$ for the current task.

For task 2 to T, we use the second task as an example to illustrate our method. After completing training, we use the first-order moment $M_2 = \{M_2^1, M_2^2,\dots, M_2^l, \dots, M_2^L\}$ obtained from the second task to calculate the component that is orthogonal to the previously gradient space:
\begin{equation}
  \label{eq:grad_proj}
  \hat{M_2^l}=M_2^l - \mathcal{S}^l(\mathcal{S}^l)^{\top}M_2^l = M_2^l - M_{2,Proj}^l
\end{equation}
We perform SVD decomposition on the first-order moment of each layer, obtaining $\hat{M}_2^l = U_2^l \Sigma_2^l {V_2^l}^{\top}$. Then we apply the updated constraints and the approximation threshold $\epsilon_{t}^l$ to perform a k-rank approximation:
\begin{equation}
  \label{eq:secondtask_k}
  \|(\hat{M}_2^l)_k\|_F^2+\|\hat{M}_{2,Proj}^l\|_F^2 \geq \epsilon_t^l\|\hat{M}_2^l\|_F^2
\end{equation}
Finally, we update the gradient space as follows:
\begin{equation}
    \label{eq:update_S}
    \mathcal{S} = [\mathcal{S}, u_{2,1}^l, u_{2,2}^l,\dots,u_{2,k}^l]
\end{equation}

As the number of tasks grows, the gradient space expands, increasing its dimensionality. To regulate this, we impose constraints by truncating smaller singular values, ensuring the gradient space remains fixed in size. This is achieved by selectively replacing gradient vectors in the shared space according to their singular values.

\subsection{Low Rank Projection Optimization}
\label{low-rank-projection-optimization}
In this section, we leverage the gradient shared space to project the training parameters effectively. Our training parameters consist of both LoRA and the full-rank parameters. The core idea behind low-rank projection is to reduce redundant information by constraining updates within the low-rank gradient space, ensuring learning focuses on critical direction updates. This approach mitigates overfitting and improves the model's generalization ability in high-dimensional data, resulting in a more stable training process, while maintaining fine-tuning efficiency.

Specifically, for LoRA parameters, the projection is applied to parameter $A$, which is projected into the gradient shared space. Given the gradient $G_{A,l}\in \mathbb{R}^{m \times n}$ of parameter $A$ and the gradient space $\mathcal{S}_{t-1}^{A,l}$:
\begin{equation}
  \label{eq:lora_proj}
  G_{A,l}^{'} = G_{A,l} - \mathcal{S}_{t-1}^{A,l} (\mathcal{S}_{t-1}^{A,l})^{\top} G_{A,l}
\end{equation}
For full-rank parameters, following \citet{DBLP:conf/icml/Zhao0CWAT24}, we apply low-rank updates during Adam optimization rather than full-rank updates. Since full-parameter training introduces additional memory overhead and given that parameter gradients tend to exhibit a low-rank structure over the course of training, it is essential to preserve their low-rank nature as much as possible throughout the optimization. Given a full-rank parameter gradient $G_{t,l} \in \mathbb{R}^{m \times n}$, we decompose it into a low-rank structure using $G_{t,l} = U_l \sum_l V_l^{\top}$, then we select first k vectors $U_{l,k}$ and $ V_{l,k}$, and project them into $G_{t,l}$ as follows:
\begin{equation}
  \label{eq:Gt_svd}
  G_{t,l}^{'} = U_{l,k}^{\top} G_{t,l} V_{l,k}
\end{equation}
The original gradient information is compressed by projecting $G_{t,l}$ into a low-rank representation $G_{t,l}^{'}$. This reduces the dimensionality of the data while preserving its most significant features. Then $G_{t,l}^{'}$ is projected into gradient space $\mathcal{S}_{t-1}^l$ as follows: 
\begin{equation}
  \label{eq:Gt_svd0}
  P_{t,l} = G_{t,l}^{'} - \mathcal{S}_{t-1}^l (\mathcal{S}_{t-1}^l)^{\top} G_{t,l}^{'}
\end{equation}
The projected gradient $G_{t,l}^{'}$ of LoRA and the low-rank projected gradient $P_{t,l}$ are then optimized by Adam:
\begin{align}
  M_{t,l} &= \beta_1 M_{t-1,l} + (1-\beta_1) P_{t,l} \label{eq:Gt_svd1}\\
  V_{t,l} &= \beta_2 V_{t-1,l} + (1-\beta_2) P_{t,l}^2 \label{eq:Gt_svd2}\\
  P_{t,l}^{'} &= M_{t,l}/\sqrt{V_{t,l}+\epsilon} \label{eq:Gt_svd3}
\end{align}
Finally, the low-rank projected gradient is scaled back to the original gradient dimension:
\begin{align}
  \hat{G_{t,l}} &= \alpha U_{l,k} P_{t,l}^{'} V_{l,k}^{\top} \label{eq:proj_back} \\
  W_{t,l} &\leftarrow W_{t-1,l} + \eta\hat{G_{t,l}} \label{eq:weight_update}
\end{align}
where $\alpha$ is the scaling factor and $\eta$ is the learning rate. LoRA gradients do not require dimensional expansion and directly update the weights with Equation~\ref{eq:weight_update}. However, frequent low-rank operations can introduce additional computational overhead. Therefore, we minimize the low-rank operations for full-rank parameters by updating them at fixed intervals. 
Simultaneously, the projection process in Equation~\ref{eq:Gt_svd} is simplified by projecting the gradients into a subspace, denoted as $G_{t,l}^{'} = U_{l,k}^{\top} G_{t,l}$.

\begin{table*}
    \centering
    \begin{tabular}{lcccccccc}
        \toprule
        & \multicolumn{4}{c}{\textbf{Standard CL Benchmark}} & \multicolumn{4}{c}{\textbf{Large Number of Tasks}} \\
        &\textbf{Order-1 }& \textbf{Order-2 }& \textbf{Order-3} & \textbf{Avg} & \textbf{Order-4} & \textbf{Order-5} & \textbf{Order-6} & \textbf{Avg} \\
        \midrule
        ProgPrompt & 75.2 & 75.1 & 75.1 & 75.1 & 78.3 & 77.9 & 77.9 & 78.0 \\
        PerTaskFT & 70.0 & 70.0 & 70.0 & 70.0 & 78.1 & 78.1 & 78.1 & 78.1 \\
        MTL & 80.0 & 80.0 & 80.0 & 80.0 & 76.5 & 76.5 & 76.5 & 76.5 \\
        \midrule
        SeqFT & 18.9 & 24.9 & 41.7 & 28.5 & 7.5 & 7.4 & 7.5 & 7.4 \\
        SeqLoRA & 44.6 & 32.7 & 53.7 & 43.7 & 2.0 & 1.9 & 1.6 & 1.8 \\
        IncLoRA & 66.0 & 64.9 & 68.3 & 66.4 & 54.7 & 53.2 & 62.2 & 56.7 \\
        Replay & 55.2 & 56.9 & 61.3 & 57.8 & 44.5 & 46.5 & 45.1 & 45.4 \\
        \midrule
        EWC & 48.7 & 47.7 & 54.5 & 50.3 & 46.9 & 45.6 & 45.6 & 46.0 \\
        LwF & 50.2 & 52.0 & 64.3 & 55.5 & 49.9 & 50.5 & 49.5 & 49.9 \\
        L2P & 60.3 & 61.7 & 61.1 & 61.0 & 56.9 & 56.9 & 56.1 & 56.6 \\
        LFPT5 & 65.3 & 68.0 & 71.5 & 68.3 & 70.0 & 73.0 & 73.8 & 72.3 \\
        \midrule
        O-LoRA & 75.4 & 75.7 & 76.3 & 75.8 & 72.3 & 64.8 & 71.6 & 69.6 \\
        MIGU & 77.1	& 77.0 & 75.6 & 76.6 & 67.3	& 68.5 & 74.2 & 70.0 \\
        N-LoRA & 79.2 & 78.4 & 78.8 & 78.8 & 73.6 & 70.3 & 73.2 & 72.4 \\
        \textbf{GORP}  & \textbf{79.7} & \textbf{79.9} & \textbf{79.7} & \textbf{79.8} & \textbf{76.1} & \textbf{76.2} & \textbf{75.6} & \textbf{76.0} \\
        \bottomrule
    \end{tabular}
    \caption{Performance comparison of different methods using the T5 model on Standard CL Benchmark and Large Number of Tasks. The average accuracy after training on the final task is reported.}
    \label{tab:performance_comparison}
\end{table*}

\section{Experiments}
In this section, we present the experimental setup and evaluate the performance of the proposed GORP method across multiple tasks. The focus is on assessing the advantages of GORP in terms of model performance and adaptability, while also comparing it with existing mainstream methods.
\subsection{Experimental Setups}
\paragraph{Models and Datasets.}
To evaluate the proposed method, we employ two widely adopted language models: the encoder-decoder T5-Large model \citep{JMLR_Raffel2020} with 770M parameters and the decoder-only LLaMA2 model \citep{touvron2023llamaopenefficientfoundation} with 7B parameters. For datasets, we utilize the standard CL benchmarks \citep{NIPS2015_250cf8b5} and the large number of tasks \citep{RazdaibiedinaMH23} as our experimental datasets. The standard CL benchmarks consist of classification datasets with 4 tasks and 5 categories, while the large number of tasks dataset includes a long-sequence CL dataset with 15 tasks, comprising the GLUE benchmark \citep{wang-etal-2018-glue}, SuperGLUE benchmark \citep{NEURIPS2019_4496bf24}, and the IMDB movie reviews dataset \citep{maas-etal-2011-learning}. Following the experimental setup of \citet{DBLP:conf/iclr/QinJ22} and \citet{wang-etal-2023-orthogonal}, we shuffle the tasks in the datasets and establish three different task orders. Detailed information is provided in Appendix~\ref{sec:datasets}.
\paragraph{Evaluation Metrics.}
We evaluate the effectiveness of our GORP method from multiple perspectives using various evaluation metrics, including Average Accuracy, Backward Transfer (BWT), Parameter Orthogonality, and Gradient Orthogonality. The detailed calculation methods are provided in Appendix~\ref{sec:metrics}.
\paragraph{Baselines.}
To demonstrate the effectiveness of our method, we compare it with various CL baseline approaches, including both non-continual learning methods and non-continual learning methods.
\begin{itemize}[leftmargin=*]
    \item \textit{\textbf{Non-Continual Learning Methods}}: \textbf{MTL} (Multi-Task Learning), which involves jointly training on multiple task datasets, typically represents the upper bound of continual learning. \textbf{PerTaskFT} trains an independent model for each task, \textbf{SeqFT} \citep{10.5555/3454287.3455464} entails continual training of all parameters, \textbf{SeqLoRA} focuses on training only one LoRA, and \textbf{IncLoRA} involves training a new LoRA for each task.
    \item \textit{\textbf{Continual Learning Methods}}: \textbf{Replay} involves merging old task data to train new tasks, while \textbf{EWC} \citep{doi:10.1073/pnas.1611835114} and \textbf{LwF} \citep{8107520} adjust model parameters using regularization losses. \textbf{L2P} \citep{9878681} and \textbf{LFPT5} \citep{DBLP:conf/iclr/QinJ22} dynamically design prompts to adapt to new tasks, and \textbf{O-LoRA} \citep{wang-etal-2023-orthogonal} constrains LoRA parameters to be orthogonal in a subspace to learn new tasks. \textbf{MIGU} \citep{du-etal-2024-unlocking} considers output gradient normalization distributions to filter parameter updates, and \textbf{N-LoRA} \citep{yang-etal-2025-parameter} reduces collisions by sparsifying parameter updates.
\end{itemize}

\begin{figure*}[t]
    \centering
  \includegraphics[width=0.92\linewidth]{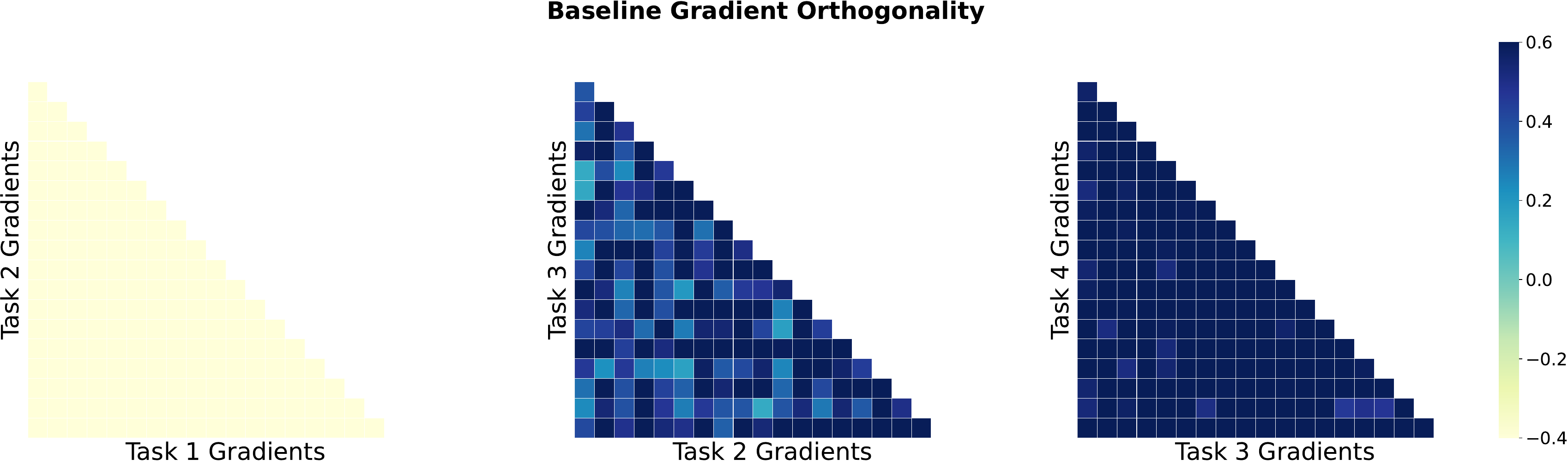}
  \includegraphics[width=0.92\linewidth]{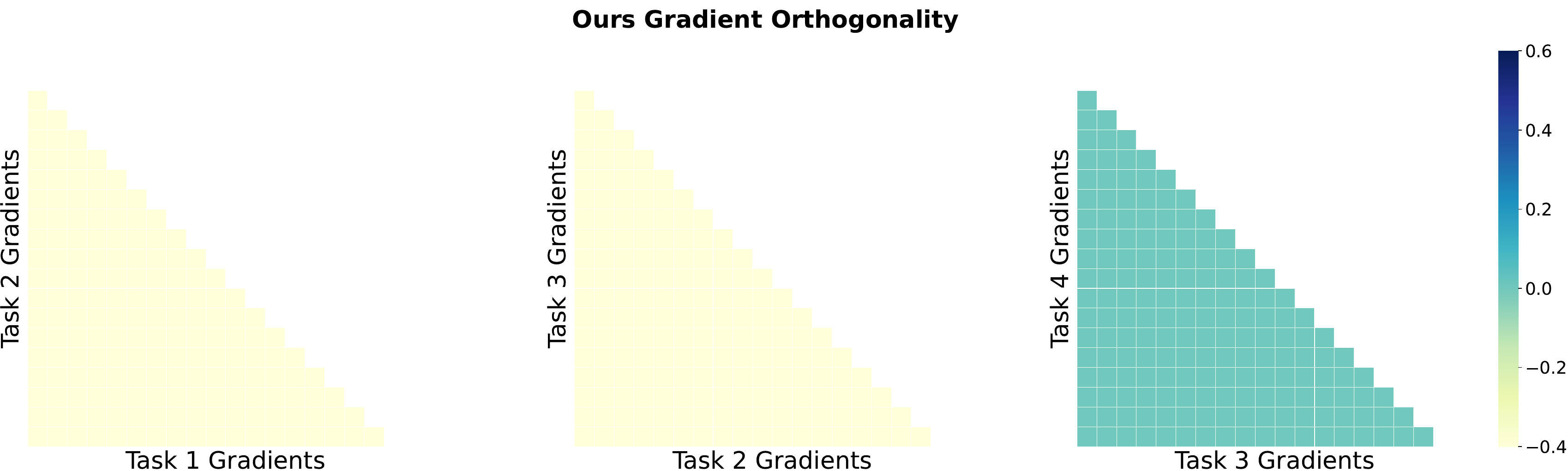}
  \caption {The visualization comparison of gradient orthogonality between Baseline and our method using the T5 model on Standard CL Benchmark. Although the first two tasks maintain orthogonality, gradient interference between parameters gradually increases as more tasks are added, while our method consistently preserves orthogonality.}
  \label{fig:orthogonality}
\end{figure*}
\begin{table}
    \centering
    \scalebox{0.92}{
    \begin{tabular}{lcccc}
    \toprule
        & \textbf{Order-1} & \textbf{Order-2} & \textbf{Order-3} & \textbf{Avg} \\
    \midrule
    O-LoRA & 76.8 & 75.7 & 75.7 & 76.1 \\
    N-LoRA & 77.2 & 77.3 & \textbf{78.4} & 77.6 \\
    \textbf{GORP} & \textbf{78.7} & \textbf{78.8} & 78.2	& \textbf{78.6} \\
    \bottomrule
    \end{tabular}}
    \caption{Performance comparison of various methods implemented on the LLaMA2-7B model, reporting average accuracy across all task orders and evaluated across multiple task orders within the Standard CL Benchmark.}
    \label{tab:performance_comparison_large}
\end{table}
\subsection{Main Results}
We compare the performance of GORP with baseline methods on two types of CL benchmarks. The experimental results across different task orders are summarized in Table~\ref{tab:performance_comparison}.
\paragraph{Performance on standard CL benchmarks.}on the T5 model, GORP demonstrates consistent superiority over all prior methods across various task sequences, achieving significant improvements on standard continual learning benchmarks. Specifically, GORP improves performance by 4\% over baseline methods while closely approaching MTL performance. As shown in Table~\ref{tab:performance_comparison_large}, GORP also significantly outperforms baseline methods on LLaMA2-7B, achieving a 2.5\% performance gain. These results highlight the effectiveness of our approach, even with larger model parameters.
\paragraph{Performance on a Large Number of Tasks.}Continual learning tasks with long sequences are generally more challenging. As shown in Table~\ref{tab:performance_comparison}, GORP consistently outperforms the baseline methods, achieving a 6.1\% performance improvement. It also surpasses other state-of-the-art methods, with GORP's performance approaching that of MTL. Additionally, GORP performs more similarly to PerTaskFT than other methods, suggesting that combining low-rank parameters with full parameters helps narrow the performance gap.

\begin{figure}[t]
  \includegraphics[width=\columnwidth]{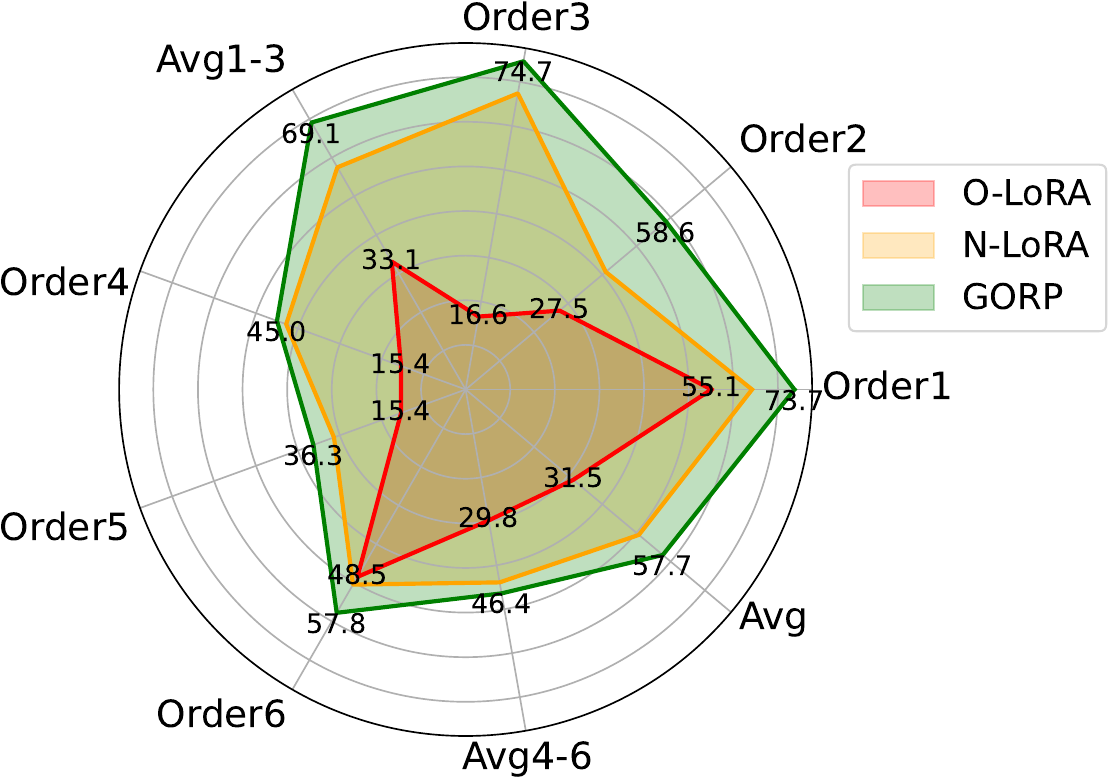}
  \caption{Performance comparison of the T5 model's generalization to unseen tasks. GORP consistently outperforms other methods across all task orders.}
  \label{fig:generalization}
\end{figure}
\begin{table}
    \centering
    \begin{tabular}{lcc}
    \toprule
        & \multicolumn{2}{c}{\textbf{BWT} (\%)} \\
        & \textbf{Avg Order 1-3} & \textbf{Avg Order 4-6} \\
    \midrule
    O-LoRA & -7.8 & -16.4 \\
    N-LoRA & -4.9 & -6.5 \\ 
    \textbf{GORP} & \textbf{-0.8} & \textbf{-4.3} \\
    \bottomrule
    \end{tabular}
    \caption{The forgetting rate comparison between the baseline and our proposed method on the T5 model, quantified using Backward Transfer (BWT) as the evaluation metric. As evidenced by the comparative results presented in the table, our method demonstrates a 7\% and 12.1\% reduction in forgetting rate compared to the baseline.}
    \label{tab:BWT}
\end{table}
\begin{table}[t]

    \begin{tabular}{lccc}
    \toprule
        & \multicolumn{3}{c}{\textbf{Method}} \\
        & \textbf{O-LoRA} & \textbf{N-LoRA} & \textbf{GORP}\\
    \midrule
    FLOPs & 68.4 & 84.3 & 0.125 \\
    ($\times 10^{12}$)& $1\times$ & $1.23\times$ & 1.8e-3$\times$ \\
    \midrule
    \multirow{2}*{Time/task} & 128.5 & 97.7 & 128.1 \\
    & $1\times$ & $0.76\times$ & $0.99\times$ \\
    \bottomrule
    \end{tabular}
    \caption{Time complexity comparison of different methods using the T5 model on Standard CL Benchmark.}
    \label{tab:time_anaysis}
\end{table}
\paragraph{Generalization of LLMs.}This part explores the generalization ability of our proposed GORP. We train on the first T-1 tasks, and test on the unseen t-th task, evaluating directly on the unseen task for comparison. As shown in Figure~\ref{fig:generalization}, although O-LoRA and its improved version, N-LoRA, outperform the pre-trained model on unseen tasks, the GORP method surpasses these comparative methods in generative ability. Across all task order configurations, GORP surpasses N-LoRA and O-LoRA, achieving average performance improvements of 7.0\% and 26.2\%, respectively. The results demonstrate the superior generative capability of GORP on unseen tasks.

\begin{figure}[t]
  \includegraphics[width=\columnwidth]{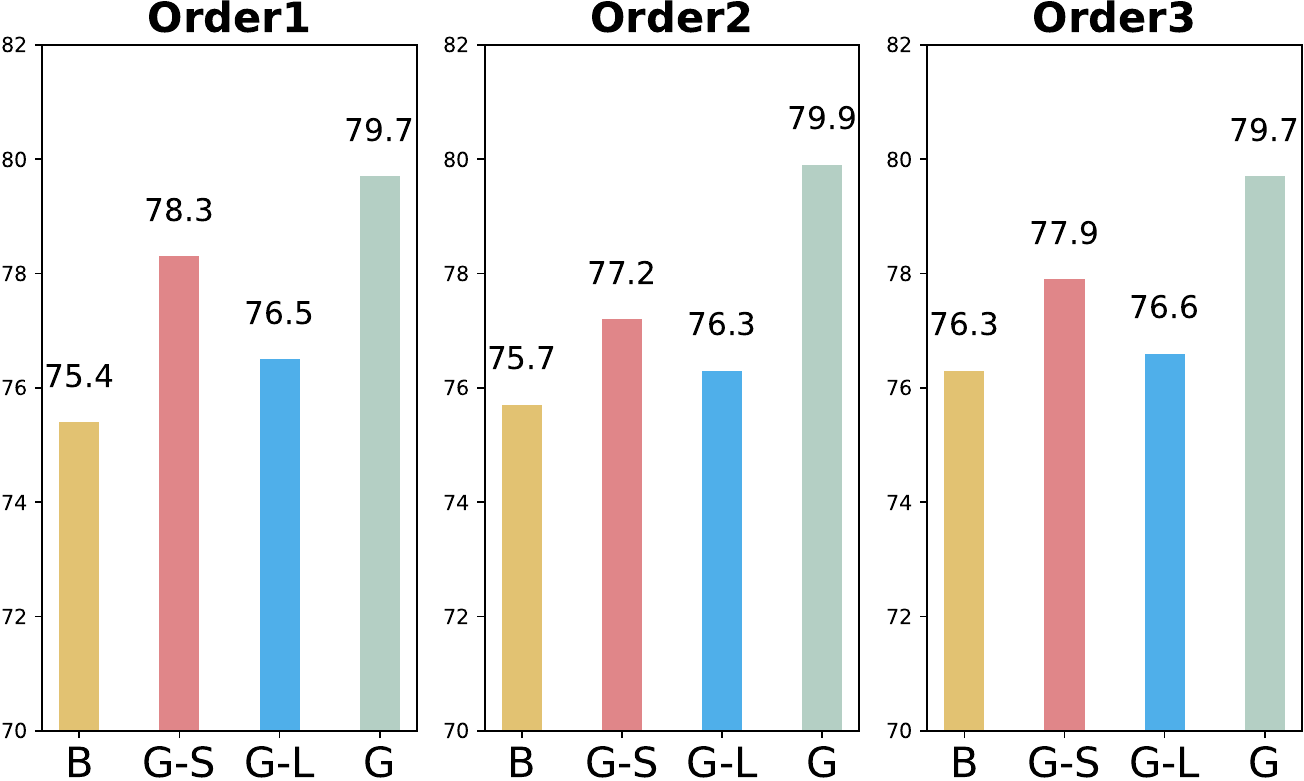}
  \caption{Ablation study of our method. B refers to the baseline method, L refers to low-rank projection for full-rank parameters, S refers to projection for LoRA, and G refers to our GORP method, which outperforms other components.}
  \label{fig:ablation}
\end{figure}

\subsection{Ablation Study}
In this section, we conduct ablation experiments to assess the contribution of each component to GORP. As shown in Figure~\ref{fig:ablation}, adding low-rank projections to LoRA improves performance by an average of 0.7\% compared to the baseline. Combining LoRA with full-rank parameters and low-rank projection results in an average improvement of 2.0\%, while the overall improvement reaches 3.9\%. The results suggest that the incorporating both full-rank and low-rank parameters produces a complementary effect. The full-rank parameters enhance model flexibility and enable finer-grained adjustments, leading to improved performance. The ablation results confirm the effectiveness of each component.

\subsection{Model Forgetting}
Forgetting is a critical challenge in continual learning. To address this, we compare the forgetting rate of GORP with baseline methods. As shown in Table~\ref{tab:BWT}, GORP achieves a forgetting rate of just 0.8\%, while baseline methods exhibit a rate of 7.8\%, representing a 7.0\% reduction. This result highlights the strong anti-forgetting capability of GORP.

\begin{figure}[t]
  \includegraphics[width=\columnwidth]{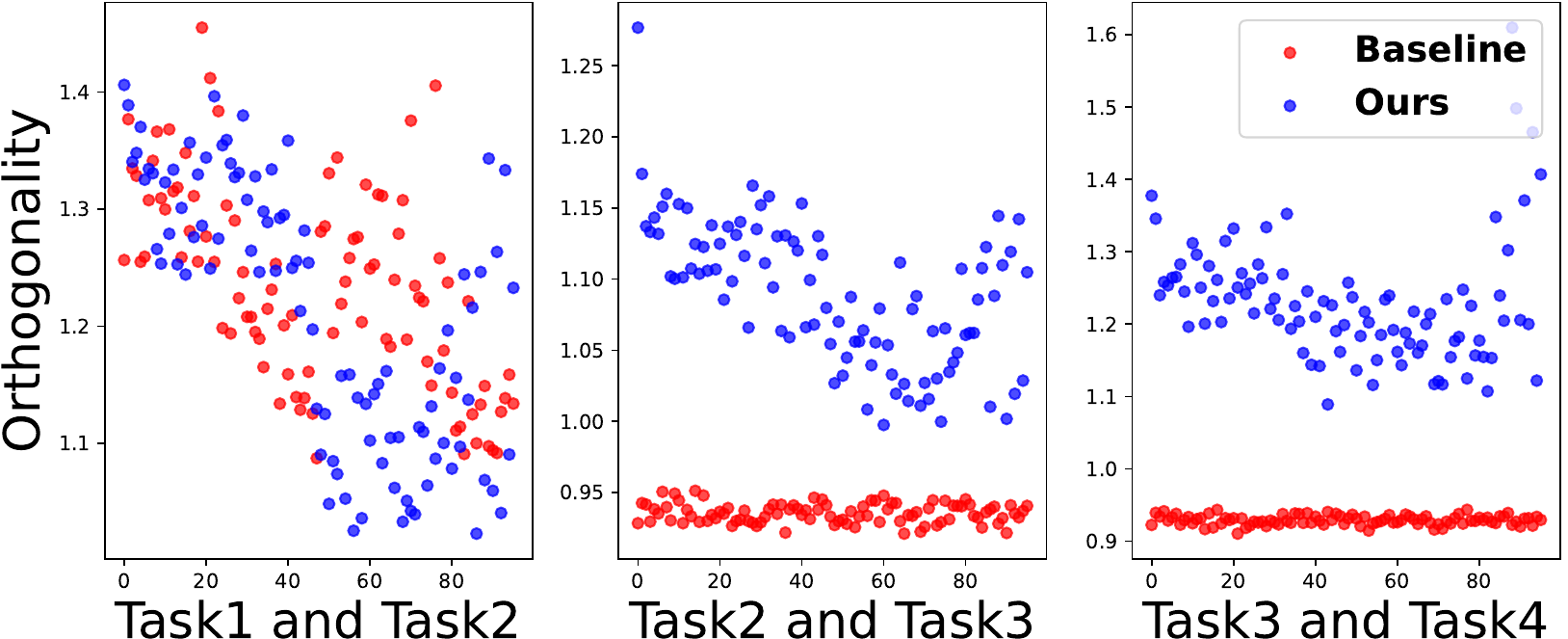}
  \caption{The visualization comparison of parameter orthogonality between baseline and our method using the T5 model. Although the parameter orthogonality of our method is higher compared to the baseline, the difference is not significant.}
  \label{fig:param}
\end{figure}

Gradient space plays a crucial role in mitigating forgetting. While O-LoRA explicitly enforces orthogonality constraints on LoRA weights, GORP applies implicit constraints to regulate gradients. We compare the updates of parameter $A$ in GORP and O-LoRA from both parameter and gradient perspectives, visualizing the weight distribution of $A$ and the orthogonality of gradient distributions. 
As shown in Figure~\ref{fig:param}, the baseline method maintains  parameter orthogonality throughout. Although GORP exhibits slightly weaker parameter orthogonality, the difference is minimal. However, GORP demonstrates highly stable gradient orthogonality in Figure~\ref{fig:orthogonality}, enabling better gradient direction control while allowing parameters to update within a larger space, thereby increasing their degrees of freedom.
\subsection{Time Complexity Analysis}
We present in Table~\ref{tab:time_anaysis} the floating point operations per second (FLOPs) and total running times (in seconds) of different methods on the standard CL benchmarks. Compared to O-LORA, our proposed GORP method requires nearly the same amount of time but significantly reduces computational cost. In contrast, N-LoRA reduces training time but increases computational demand. This indicates that our GORP method does not introduce significant computational delays and optimizes efficiency, making it a more resource-efficient alternative to O-LORA. While N-LoRA offers desirable speedup, it may result in higher computational burden. Therefore, GORP may be more suitable for scenarios where both time and computational resources are critical.
\section{Conclusion}
In this work, we propose GORP,  a novel training strategy that overcomes these limitations by synergistically combining full and low-rank parameters and jointly updating within a unified low-rank gradient subspace. GORP is enable to expand the search space for optimal solutions while preserving the essential properties of continual fine-tuning. Through extensive empirical evaluations, we show that GORP effectively addresses the stability-plasticity dilemma in continual learning, all while maintaining computational efficiency during the fine-tuning.

\section*{Limitations}
While GORP outperforms existing methods on continual learning benchmarks, several limitations should be considered. 
First, as task sequences expand, continuously updating task vectors within the gradient subspace becomes necessary. Therefore, effectively capturing increasing task diversity within constrained dimensional boundaries is a key challenge. 
Additionally, while GORP has shown strong performance in known continual data environments, its effectiveness in more complex real-world scenarios remains to be further validated.

\section*{Acknowledgments}
This work was partly supported by the National Key Research and Development Program of China under Grant 2024YFE0202900; the National Natural Science Foundation of China under Grant (62436001,62176020); the Joint Foundation of the Ministry of Education for Innovation team (8091B042235); and the State Key Laboratory of Rail Traffic Control and Safety (Contract No. RCS2023K006), Beijing Jiaotong University.
\bibliography{anthology,custom}

\appendix
\section{Preliminary Knowledge}
\subsection{Continual Learning Setup}
For consecutive tasks $\{T_1,T_2,\dots,T_n\}$, each task $T_t$ contains $N_t$ samples $\{x_t,y_t\}_{t=1}^{N_t}$. In the t-th task, each step will sample n training samples $\mathcal{B}_n$ from the task for training, obtain parameter weights $W_{s}^t$, and then accumulate the weights to obtain the weight of the current task $W_t=\sum_s{W_s^t}$, and integrate with the previous task weight to get $W_t^{'}=W_{t-1}^{'} + W_t$. The model is able to retain its performance on previous tasks while progressively learning new ones, thereby minimizing the forgetting of earlier tasks.
\subsection{Low-Rank Adaptation}
For a pre-trained weight $W_p \in \mathbb{R}^{m \times n}$,  LoRA freezes the pretrained parameters and updates $W_{new} = W_p+\Delta W=W_p + AB$ by training low rank parameters, where $A \in \mathbb{R}^{m \times k}$ and $B \in \mathbb{R}^{k \times n}$, and rank $k \ll min(m,n)$. 
For a linear layer, the output can be written by Equation~\ref{eq:linear}:
\begin{equation}
  \label{eq:linear}
  y = (W_p+\Delta W)x = W_px + ABx
\end{equation}
Through low rank updates, $W_{new}$  retains the capabilities of pretrained models and also improves the generalization ability on downstream tasks.

\section{Datasets and Task Details}
\label{sec:datasets}
This part presents the datasets used in the experiments, along with the data categories and their corresponding tasks. The detailed information is provided in Table~\ref{tab:datasets}. CL benchmark includes Yelp, Amazon, Dbpedia, Yahoo and Agnews, GLUE dataset includes MNLI, QQP, RTE and SST-2, and SuperGLUE includes WiC, CB, COPA, BoolQA, MultiRC and IMDB. For the large number of tasks, we select 1000 random samples for training each task and 500 samples per class for validation and testing.

We report the task sequences used for CL experiments on the T5 and LLaMA2 models in Table~\ref{tab:appendix_task_sequence}. These datasets span diverse categories, including natural language inference (NLI), sentiment classification (SC), and topic classification (TC), ensuring diverse abilities of the model’s generalization across multiple tasks. And the task instructions for different categories are shown in Table~\ref{tab:instructions}.
\begin{table*}
    \centering
    \scalebox{0.95}{
    \begin{tabularx}{\linewidth}{llllc}
    \toprule
    \textbf{Dataset Name} & \textbf{Category} & \textbf{Task} & \textbf{Domain} & \textbf{Metric}\\
    \midrule
    Yelp & CL Benchmark & Sentiment analysis & Yelp reviews & Accuracy \\
    Amazon & CL Benchmark & Sentiment analysis & Amazon reviews & Accuracy \\
    Dbpedia & CL Benchmark & Topic classification & Wikipedia & Accuracy \\
    Yahoo & CL Benchmark & Topic classification & Yahoo Q\&A & Accuracy \\
    AG News & CL Benchmark & Topic classification & News & Accuracy \\
    MNLI & GLUE & NLI & Various & Accuracy \\
    QQP & GLUE & Paragraph detection & Quora & Accuracy \\
    RTE & GLUE & NLI & News, Wikipedia & Accuracy \\
    SST-2 & GLUE & Sentiment analysis & Movie reviews & Accuracy \\
    WiC & SuperGLUE & Word sense disambiguation & Lexical databases & Accuracy \\
    CB & SuperGLUE & NLI & Various & Accuracy \\
    COPA & SuperGLUE & QA & Blogs, encyclopedia & Accuracy \\
    BoolQA & SuperGLUE & Boolean QA & Wikipedia & Accuracy \\
    MultiRC & SuperGLUE & QA & Various & Accuracy \\
    IMDB & SuperGLUE & Sentiment analysis & Movie reviews & Accuracy \\
    \bottomrule
    \end{tabularx}}
    \caption{Datasets, Categories, Domians and evaluation Metrics.}
    \label{tab:datasets}
\end{table*}
\begin{table*}
    \centering
    \begin{tabular}{lcc}
    \toprule
    \textbf{Model} & \textbf{Order} & \textbf{Task Sequence} \\
    \midrule
    T5-Large, LLaMA2 & \textbf{1} & dbpedia $\rightarrow$ amazon $\rightarrow$ yahoo $\rightarrow$ ag\\
    T5-Large, LLaMA2 &\textbf{2} &  dbpedia $\rightarrow$ amazon $\rightarrow$ ag $\rightarrow$ yahoo\\
    T5-Large, LLaMA2 & \textbf{3} & yahoo $\rightarrow$ amazon $\rightarrow$ ag $\rightarrow$ dbpedia\\
    \midrule
    T5-Large & \textbf{4} & \makecell{mnli $\rightarrow$ cb $\rightarrow$ wic $\rightarrow$ copa $\rightarrow$ qqp $\rightarrow$ boolqa $\rightarrow$ rte $\rightarrow$ imdb $\rightarrow$ \\  yelp $\rightarrow$ amazon $\rightarrow$ sst-2 $\rightarrow$ dbpedia $\rightarrow$ ag $\rightarrow$ multirc $\rightarrow$ yahoo}\\
    T5-Large & \textbf{5} & \makecell{multirc $\rightarrow$ boolqa $\rightarrow$ wic $\rightarrow$ mnli $\rightarrow$ cb $\rightarrow$ copa $\rightarrow$ qqp $\rightarrow$ rte $\rightarrow$ \\ imdb $\rightarrow$ sst-2 $\rightarrow$ dbpedia $\rightarrow$ ag $\rightarrow$ yelp $\rightarrow$ amazon $\rightarrow$ yahoo}\\
    T5-Large & \textbf{6} & \makecell{yelp $\rightarrow$ amazon $\rightarrow$ mnli $\rightarrow$ cb $\rightarrow$ copa $\rightarrow$ qqp $\rightarrow$ rte $\rightarrow$ imdb $\rightarrow$ \\ sst-2 $\rightarrow$ dbpedia $\rightarrow$ ag $\rightarrow$ yahoo $\rightarrow$ multirc $\rightarrow$ boolqa $\rightarrow$ wic}\\
    \bottomrule
    \end{tabular}
    \caption{Task sequences used for CL experiments on the T5 and LLaMA2 models.}
    \label{tab:appendix_task_sequence}
\end{table*}
\begin{table*}
    \centering
    \scalebox{0.95}{
    \begin{tabularx}{0.8\linewidth}{l >{\raggedright\arraybackslash}X }
    \toprule
    \textbf{Task} & \textbf{Instructions} \\
    \midrule
    NLI & What is the logical relationship between the "sentence 1" and the "sentence 2"? Choose one from the option. \\
    \midrule
    QQP & Whether the "first sentence" and the "second sentence" have the same meaning? Choose one from the option. \\
    \midrule
    SC & What is the sentiment of the following paragraph? Choose one from the option. \\
    \midrule
    TC & What is the topic of the following paragraph? Choose one from the option. \\
    \midrule
    BoolQA & According to the following passage, is the question true or false? Choose one from the option. \\
    \midrule
    MultiRC & According to the following passage and question, is the candidate answer true or false? Choose one from the option. \\
    \midrule
    WiC & Given a word and two sentences, whether the word is used with the same sense in both sentences? Choose one from the option. \\
    \bottomrule
    \end{tabularx}}
    \caption{Instructions for different tasks.}
    \label{tab:instructions}
\end{table*}

\section{Evaluation Metrics}
\label{sec:metrics}
Let $a_{i,j}$ be the test accuracy of the $i$-th task after training on the $j$-th task. $A_i$ denotes the A matrix of LoRA, and $G_{A,i}$ denotes the gradient of A matrix on the $i$-th task. We evaluate the model using the following metrics:
\begin{itemize}[leftmargin=*]
    \item \textbf{Average Accuracy (ACC)}: The average accuracy of all tasks after training on the last task:
    \begin{equation}
        ACC = \frac{1}{T}\sum_{i=1}^T{a_i,T}
    \end{equation}
    \item \textbf{Backward Transfer (BWT)}: The average forgetting of all tasks after training on the last tasks:
    \begin{equation}
        BWT = \frac{1}{T-1}\sum_{i=1}^{T-1}{a_{i,T} - a_{i,i}}
    \end{equation}
    \item \textbf{Parameter Orthogonality (PO)}: We use this metric to quantify the orthogonal overlap between $A_i$ and $A_j$, for the reason that O-LoRA use $A$ to capture gradient subspaces of previous tasks. The metric is  calculated as:
    \begin{equation}
        PO_{i,j} = \|A_i^{\top}A_j\|^2
    \end{equation}
    \item \textbf{Gradient Orthogonality (GO)}: We use this metric to quantify the orthogonal overlap between $G_{A,i}$ and $G_{A,j}$, showing the difference between the gradient space and the parameter space, calculated as:
    \begin{equation}
        GO_{i,j} = \|G_{A,i}^{\top}G_{A,j}\|^2
    \end{equation}
\end{itemize}

\begin{table*}
\centering
\begin{tabular}{lllccc}
\toprule
\textbf{Category} & \textbf{Dataset} & \textbf{Source} & \textbf{Avg len} & \textbf{Metric} & \textbf{Language} \\
\midrule
\multirow{3}*{Domain-specific} &ScienceQA & Science & 210 & Accuracy & English \\
& FOMC & Finance & 51 & Accuracy & English \\
& MeetingBank & Meeting & 2853 & ROUGE-L & English \\
\midrule
\multirow{2}*{Multi-lingual} & C-STANCE & Social media & 127 & Accuracy & Chinese \\
& 20Minuten & News & 382 & SARI & German \\
\midrule
Code Completion & Py150 & Github & 422 & Edit Similarity & Python \\
\midrule
\multirow{2}*{Mathematical Reasoning} & NumGLUE-cm & Math & 32 & Accuracy & English \\
& NumGLUE-ds & Math & 21 & Accuracy & English \\
\bottomrule
\end{tabular}
\caption{The overview of dataset statistics in TRACE, where 'SARI' is a score that is specific to evaluating simplification tasks.}
\label{tab:appendix-datasets}
\end{table*}

\section{Implementation Details}
We adapted the code-base from O-LoRA \citep{wang-etal-2023-orthogonal}. And our improved version of the code is available in the supplementary meterial and will be released upon acceptance. All experiments were conducted on the machine with 8 NVIDIA L20 and were implemented with Deepspeed. 

For the T5 model, we employed LoRA to replace the SelfAttention layers and full-rank parameter trainings for the EncDecAttention layers. For all orders, we trained the models with one epoch, a constant learning rate 1e-03 for LoRA and 1e-05 (1e-04 for Order 4 to 6) for full-rank parameters, rank $8$ for LoRA and rank $8$ for full-rank parameters, a training batch size of $8$ per device, a evaluation batch size of $64$ per device, and a weight decay rate of 0, a value $0.05$ of $\lambda$. We set different scale factors for order 1 to 6. For order 1 to 3, we set scale factor $1$ and $0.25$ for order 4 to 6. In our method, the low-rank updates are interval, and we set the update gap $10$.

For the LLaMA2 model, we employed LoRA to replace the Self-attn layers and full-rank parameter trainings for the MLP Gate layers. For order 1 to 3, we trained the models with one epoch, a constant learning rate 2e-04 for LoRA and 1e-06 for full-rank parameters, rank $8$ for LoRA and rank $8$ for full-rank parameters, a training batch size of $1$ per device, a evaluation batch size of $4$ per device, and a weight decay rate of 0, a value $0$ of $\lambda$. We set scale factor $0.25$ for order 1 to 3 and the value $20$ of the interval gap for low-rank updates.


\begin{table}
    \centering
    \begin{tabular}{cc}
        \toprule
        \textbf{k-dim} & \textbf{Order 1} \\
        \midrule
        4 & 76.5 \\
        \textbf{8} & \textbf{79.7} \\
        16 & 79.0 \\
        32 & 77.9 \\
        64 & 77.4 \\
        \bottomrule
    \end{tabular}
    \caption{Different k values for full-rank parameters on final results for the order 1 tasks on the T5 model with Standard CL Benchmark from the standard continual learning benchmark.}
    \label{tab:appdix-k-values}
\end{table}

\begin{table}
    \centering
    \begin{tabular}{ccc}
        \toprule
        \multirow{2}*{\textbf{k-dim}} & \textbf{Yahoo} & \textbf{AG News}\\
        ~ & \textbf{(Ten-class)} & \textbf{(Four-class)} \\
        \midrule
        4 & 70.2 & 91.1\\
        \textbf{8} & \textbf{71.3} & \textbf{91.5}\\
        16 & 71.2 & 91.4\\
        32 & 70.9 & 91.4\\
        64 & 70.9 & 91.5\\
        \bottomrule
    \end{tabular}
    \caption{Performance comparison under different task complexity with varying k values, illustrated using the T5 model on the Yahoo (10-class) and AG News (4-class) tasks.}
    \label{tab:appendix-task-complexity}
\end{table}

\begin{table*}
    \centering
    \begin{tabular}{lc}
    \toprule
    \textbf{Model} & \textbf{Task Sequence} \\
    \midrule
    LLaMA2 & \makecell{c-stance $\rightarrow$ fomc $\rightarrow$ meetingbank $\rightarrow$  py150 $\rightarrow$ scienceqa \\
    $\rightarrow$ numglue-cm $\rightarrow$ numglue-ds $\rightarrow$ 20minuten}\\
    \bottomrule
    \end{tabular}
    \caption{Task sequence used for TRACE on the LLaMA2 model.}
    \label{tab:appendix_trace_sequence}
\end{table*}

\begin{table}
    \centering
    \begin{tabular}{lcccc}
        \toprule 
         & \multicolumn{4}{c}{\textbf{\#Data}} \\
        \cmidrule(r){2-5}
        \multirow{2}*{\textbf{Method}} & \multicolumn{2}{c}{\textbf{500}} & \multicolumn{2}{c}{\textbf{5000}} \\
        \cmidrule(r){2-3} \cmidrule(r){4-5}
         & \textbf{Avg} & \textbf{BWT}(\%) & \textbf{Avg} &\textbf{BWT}(\%) \\
        \midrule
        O-LoRA & 39.5 & -4.5 & 43.8 & -4.3 \\
        \textbf{GORP} & \textbf{47.3} & \textbf{-1.0} & \textbf{50.4} & \textbf{-0.7} \\
        \bottomrule
    \end{tabular}
    \caption{Comparison of between the baseline and GORP method on the LLaMA2 model.}
    \label{tab:appendix-trace}
\end{table}

\section{Extended Explanations and Results}
\subsection{Impacts of Params and Task Complexity}
To investigate the influence of the rank parameter (k) on the performance of low-rank gradients, we conducted comparative experiments on the T5 model using a standard continual learning benchmark. As an example, Table~\ref{tab:appdix-k-values} shows the impact of varying k values on the final results for the order 1. From the data, we observe that the rank of $k=8$ yields superior performance compared to other values. This finding indicates that k=8 represents an effective trade-off, enabling robust learning of high-dimensional features without exceeding the parameter constraints imposed by the low-rank factorization.

In addition, we analyze how different k values affect the results with different task complexity, in order to examine the connection between task complexity and the chosen k value. The results of this analysis are shown in the table~\ref{tab:appendix-task-complexity}.

The experimental results demonstrate that first-order performance peaks at $k=8$ and $k=16$ as k increases. Notably, tasks with varying data complexity exhibit distinct trends: the Yahoo dataset achieves optimal performance at $k=8$, while AG News results remain stable across different k values. Considering the overall empirical trends and performance trade-offs across tasks of differing complexity, we select $k=8$ as the optimal rank for full-rank parameters.

\subsection{Consideration of Computational Overhead}
We argue that performing low-rank operations on full-rank parameters during gradient updates introduces additional computational overhead, particularly when such operations are executed frequently. To mitigate this, in Section~\ref{low-rank-projection-optimization} and Algorithm~\ref{algorithm}, we adopt a sparse low-rank update strategy, where low-rank decomposition is applied at fixed intervals rather than at every optimization step. This approach substantially reduces the number of required low-rank operations. Between these intervals, we reuse the previously computed low-rank matrix, further minimizing computational costs. Given our experimental configuration, the computational burden induced by these intermittent low-rank operations remains negligible.

\subsection{Complex Scenarios Results}
To better address the challenges posed by increasingly complex environments, we introduce TRACE~\citep{wang2023tracecomprehensivebenchmarkcontinual}, a continual learning (CL) benchmark specifically designed for large language models (LLMs). This benchmark integrates eight distinct datasets, covering a range of competencies including multiple-choice QA, multilingual understanding, code generation, and mathematical reasoning, as detailed in Table~\ref{tab:appendix-datasets}. TRACE is distinguished by its significantly enhanced diversity and the deliberate inclusion of unrelated tasks.

\paragraph{Performance on TRACE.} We compare the baseline and GORP on the TRACE dataset using the LLaMA2 model. As detailed in Table~\ref{tab:appendix-trace}, GORP achieves superior results compared to O-LoRA across the entire TRACE benchmark. Specifically, GORP achieves a 7.8\% and 6.6\% boost in performance and a 3.4\% and 3.6\% lower forgetting rate for the 500-data and 5000-data settings, respectively. 
This underscores its enhanced adaptability and efficacy in complex continual learning tasks and demonstrates its continued effectiveness on unrelated tasks. Our approach thus offers a more comprehensive and expansive evaluation framework than those previously considered, encompassing a broader array of data types.

\end{document}